\documentclass[runningheads]{llncs}

\usepackage{amsmath,amssymb,amsfonts}
\usepackage[T1]{fontenc}
\usepackage{makecell}
\usepackage{graphicx}
\usepackage{multirow}
\usepackage{float}  
\usepackage{placeins}  
\usepackage[utf8]{inputenc}
\begin{document}
%
%
\title{Self-Supervised Ultrasound-Video Segmentation with Feature Prediction and 3D Localised Loss}%
\titlerunning{SSL for US-Video Segmentation with 3D Localisation Loss}
%
\author{Edward Ellis\inst{1}
\and Robert Mendel\inst{2}
\and Andrew Bulpitt\inst{1} 
\and Nasim Parsa\inst{2}
\and Michael F Byrne\inst{2}
\and Sharib Ali\inst{1}} 

\authorrunning{E. Ellis et al.}
\institute{School of Computer Science, University of Leeds, Leeds, UK, LS2 9JT 
\and Dova Health Intelligence Inc., Vancouver, Canada, V6B 2W9}

%
\maketitle              
\begin{abstract}
%
Acquiring and annotating large datasets in ultrasound imaging is challenging due to low contrast, high noise, and susceptibility to artefacts. This process requires significant time and clinical expertise. Self-supervised learning (SSL) offers a promising solution by leveraging unlabelled data to learn useful representations, enabling improved segmentation performance when annotated data is limited. Recent state-of-the-art developments in SSL for video data include V-JEPA, a framework solely based on feature prediction, avoiding pixel level reconstruction or negative samples. We hypothesise that V-JEPA is well-suited to ultrasound imaging,  as it is less sensitive to noisy pixel-level detail while effectively leveraging temporal information. To the best of our knowledge, this is the first study to adopt V-JEPA for ultrasound video data. Similar to other patch-based masking SSL techniques such as VideoMAE, V-JEPA is well-suited to ViT-based models. However, ViTs can underperform on small medical datasets due to lack of inductive biases, limited spatial locality and absence of hierarchical feature learning. To improve locality understanding, we propose a novel 3D localisation auxiliary task to improve locality in ViT representations during V-JEPA pre-training. Our results show V-JEPA with our auxiliary task improves segmentation performance significantly across various frozen encoder configurations, with gains up to 3.4\% using 100\% and up to 8.35\% using only 10\% of the training data.
\end{abstract}
\section{Introduction}
Ultrasound (US) imaging is widely used in clinical practice as a low-cost, non-invasive, and portable alternative to CT and MRI. However, building large annotated US datasets is challenging due to high noise, low contrast, and common artefacts like reverberation, acoustic shadowing, and mirror imaging \cite{quien_ultrasound_2018}. These factors make US interpretation complex, requiring significant time and expertise, often resulting in high inter-operator variability \cite{brattain_machine_2018}. US videos support clinical understanding by allowing clinicians to anticipate and identify anatomical structures and pathology across frames, offering contextual information that single-image datasets lack. Video data also aligns more closely with real-world clinical acquisition workflows.


To assist clinician interpretation of US, self-supervised learning (SSL) offers a promising solution~\cite{jiao_self-supervised_2020}. SSL leverages unlabelled data to learn useful representations, improving downstream segmentation with limited labels. SSL has often been applied to US images \cite{fu_anatomy-aware_2022,ellis_self-supervised_2025,jiao_usfm_2024} and recently to US video data \cite{jiao_self-supervised_2020,chen_contrastive_2023,e_lamoureux_segmenting_2023}, harnessing spatial and temporal information. However, many SSL approaches in US imaging present domain-specific pretext learning strategies to improve representation learning, in a contrastive \cite{fu_anatomy-aware_2022,gomez_fetal_2025} or generative SSL framework~\cite{e_lamoureux_segmenting_2023,szijarto_masked_2024}. Contrastive learning often requires many negative samples, risking degraded representations from false negatives and demanding large batches or memory banks \cite{liu_self-supervised_2023}. Generative SSL on the other hand often emphasises pixel-level reconstruction, increasing susceptibility to noise, with less emphasis on learning high-level structures \cite{liu_self-supervised_2023}. The Video Joint Embedding Prediction Architecture (V-JEPA) framework, however, addresses these limitations by avoiding both negative sampling and pixel reconstruction, and instead focussing on abstract representations through masked latent feature prediction. V-JEPA have shown state-of-the-art performance on several natural scene video datasets for classification tasks and demonstrates particular benefit in motion understanding~\cite{bardes_adrien_revisiting_2024}.

SSL methods that utilise masked patches, such as VideoMAE \cite{tong_videomae_2022} or V-JEPA \cite{bardes_adrien_revisiting_2024}, often favour transformer-based models, as positional embeddings provide essential spatial and temporal context for predicting masked regions during pre-training. This is evident in both methods offering pre-trained weights exclusively for Vision Transformer (ViT) models. This poses a challenge for medical imaging uses, with ViTs benefitting from large datasets, performance can suffer in small data scenarios, often performing worse than the convolutional neural network~\cite{zhu_understanding_2023}. This drop in performance with ViTs is often attributed to the lack of inductive bias, limited inherent locality and absence of hierarchical feature learning \cite{dosovitskiy_image_2020}. Several works have proposed techniques to help mitigate these issues. For example, Akkaya \textit{et al.}~\cite{akkaya_enhancing_2024} introduced the LIFE module to incorporate local inductive bias by adding depth-wise separable convolutional layers, which provided local context to ViT embeddings. Liu \textit{et al.}~\cite{liu_efficient_2021} introduced a dense localisation auxiliary task to encourage the ViT to learn spatial relations within an image. In addition architectures like the pyramid vision transformer \cite{wang_pyramid_2021} and SWIN transformer \cite{liu_swin_2021} improve hierarchical feature learning through progressive downsampling and local self-attention mechanisms. 

We hypothesise that V-JEPA is well suited for ultrasound image segmentation, as its avoidance of pixel-level reconstruction mitigates sensitivity to noise and low contrast problems in US data. Its ability to model spatial and temporal dynamics coherently can help to distinguish between anatomy and artefacts across frames. However, while V-JEPA is well suited to ViT-based models, its performance can suffer in low-data regimes~\cite{zhu_understanding_2023}. To address this, we propose a novel 3D localisation auxiliary task that enhances spatial and temporal sensitivity during V-JEPA pretraining, improving ViT's inherent locality limitation in limited data setting. Our approach is model-agnostic, enabling the use of pre-trained weights for domain-specific pretraining without modifying the ViT architectures. Key contributions of our work include:
\begin{enumerate}
    \item Adopting state-of-the-art V-JEPA SSL framework for medical video US image segmentation, in our case cardiac US videos
    \item Addressing the inherent locality limitation of ViT-based V-JEPA on small US video data, by improving its locality through a novel learnable 3D localisation auxiliary task.
    \item Comprehensive evaluation of video-based SSL techniques on cardiac US videos, including variable dataset sizes.
    \end{enumerate}

\section{Method}
\begin{figure*}[t!]
  \centering
  \centerline{\includegraphics[width=1.0\textwidth]{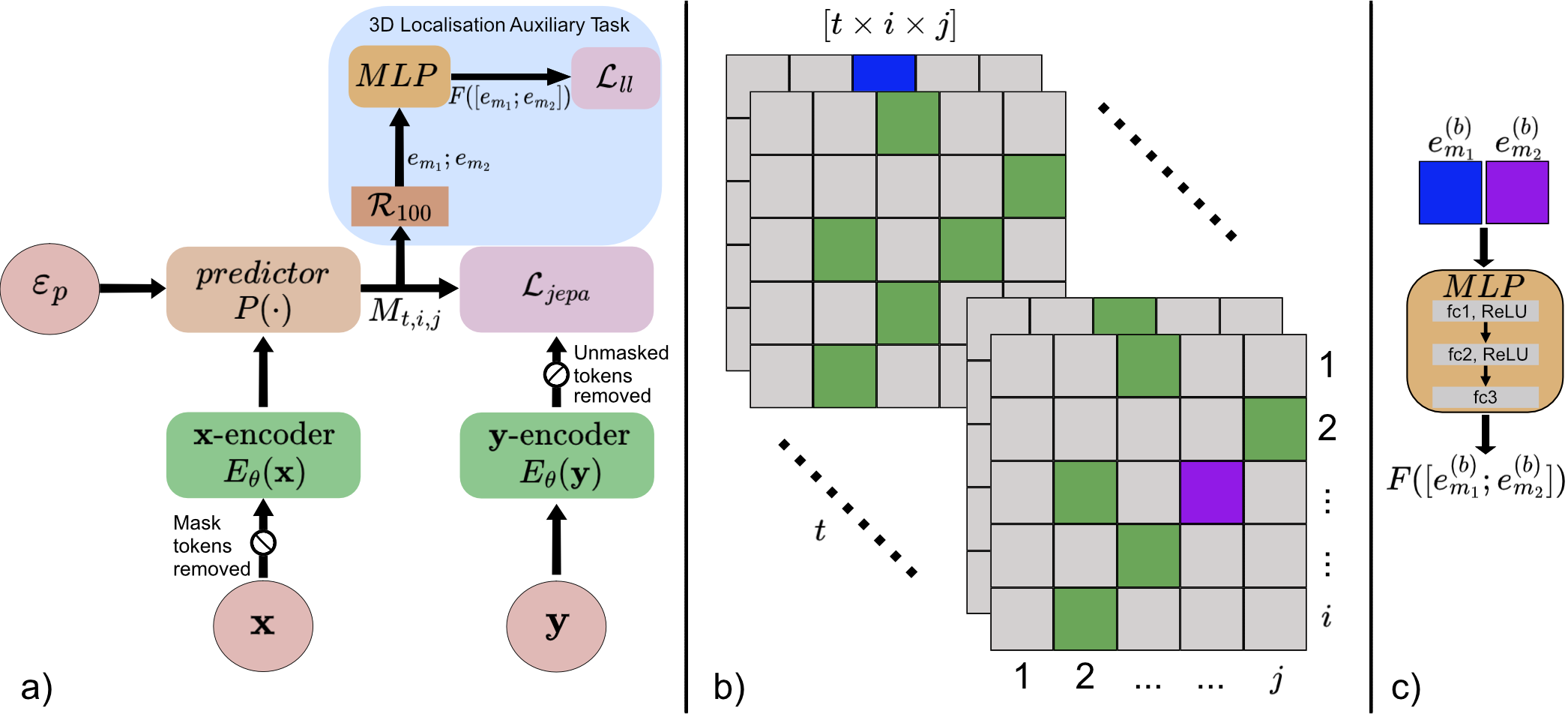}}
  \caption{Block Diagram of our 3D localisation auxiliary task incorporated in the V-JEPA SSL framework. Our auxiliary task takes a random pair of patch embeddings, ($e_{m_1}, e_{m_2}$), from the predictor and predicts the relative temporal, vertical and horizontal distances between the samples. (a) presents the general SSL framework of V-JEPA with our localisation task. (b) demonstrates the relative localisation between a sample pair (in blue and purple), sampled from the predicted masked areas, $M_{t,i,j}$ (in green). (c) shows the relative localisation prediction, ($F([e_{m_1}^{(b)};e_{m_2}^{(b)}])$), from a simple multilayer perceptron (MLP) network between the concatenated sample pair ($e_{m_1}^b; e_{m_2}^b$).}
    \label{fig:SSL_Framework}
\end{figure*}

The overall SSL approach proposed is outlined in Figure \ref{fig:SSL_Framework}. Below we detail V-JEPA for video segmentation and 3D localisation loss integrated in the V-JEPA framework. 
%
\subsection{V-JEPA}
V-JEPA builds upon the joint-embedding predictive architecture (JEPA) \cite{lecun_path_2022}. JEPA learns by predicting the representations of an input $\bf{y}$ from another input $\bf{x}$, conditioned on the transformation/corruption between $\bf{x}$ and $\bf{y}$. In practice this corruption is implemented via masking, which is contextualised for the network through positional embeddings, $\epsilon_p$. We define $\epsilon_p \leftarrow \Delta \bf{y}$, where $\Delta \bf{y}$ denotes the spatio-temporal positions of the masked regions of $\mathbf{y}$. The $\bf{x}$-encoder, $E_\theta(\bf{x})$, is trained on a masked video sequence, outputting an embedding vector for each 'visible' spatio-temporal token. The positional embeddings and output of $\bf{x}$-encoder are passed to a predictor network, $P_\phi(\cdot)$, to predict the representation of masked tokens. $P_\phi(\cdot)$ is trained simultaneously to $E_\theta(\bf{x})$.


V-JEPA effectively aims to minimise the $L_1$ loss between predicted and target representations of masked regions (see Eq. \ref{eq:jepa_loss}). Target representations are obtained by using the same encoder on the complete video clip sample, $E_\theta(\bf{y})$, with unmasked areas removed. This encoder is not trained (stop gradient $sg$ is used), and updated through the exponential moving average ($\overline{E}_{\theta}(\cdot))
$ of $E_\theta(\bf{x})$.

\begin{equation}
\mathcal{L}_{\text{jepa}} =  \ \left\| P_{\phi}\left( E_{\theta}(\bf{x}), \Delta y \right) - \operatorname{sg}\left(\overline{E}_{\theta}(\bf{y}) \right) \right\|_1
\label{eq:jepa_loss}
\end{equation}

\subsection{3D Localisation Auxiliary task}
\label{Localisation task}
Our localisation task aims to improve spatial understanding during pre-training. We add this task to the outputs of the predictor. Initially, a video clip of $T$ frames and spatial resolution of $H \times W$ is tokenised, each token of shape $2 \times 16 \times 16$. When encoded, this results in a total embedding space of shape $t \times i \times j$ spatio-temporal patch embeddings, $P_{t,i,j}$, where $t$ is the number of tubelets and $i$ and $j$ are the number of spatial tokens. Masked patch embeddings, $M_{t,i,j}$ (predicted patches) are a subset of these tokens, i.e $M_{t,i,j} \subset P_{t,i,j}$. A random subset of 100 concatenated token embedding pairs, denoted $\mathcal{R}_{100}$, is sampled from $M_{t,i,j}$, $\mathcal{R}_{100} \subset M_{t,i,j}$. We compute the 3D normalised target relative translation offset ($\Delta_{m_1,m_2}^{(b)}$) between each randomly sampled embedding pair $e_{m_1}^{(b)}$ and $e_{m_2}^{(b)}$, where $m_1$ and $m_2$ correspond to the temporal ($t$)
and spatial ($i,j$) locations of each embedding. Here $b$ indexes a sample from batch $B$. This provides a ground truth. 


\begin{equation}
\Delta_{m_1,m_2}^{(b)} = \left( \frac{t_1 - t_2}{t},\ \frac{i_1 - i_2}{i},\ \frac{j_1 - j_2}{j} \right), ~i, j, t \in [-1, 1]
\end{equation}


The sampled embedding pair is concatenated and passed as input to a small MLP, $F(\cdot)$, composed of three fully connected layers. This MLP predicts the relative translation offset, denoted $F([e_{m_1}^{(b)};e_{m_2}^{(b)}])$. Our local loss, $L_{ll}$, computes the mean squared error between the predicted and ground truth relative translation offset (see Eq. \ref{eqn:localloss}). The overall loss is computed as the average over all pairs in $\mathcal{R}_{100}$, per batch ($B$), $N_{total} = 100 \times B$. $\mathcal{R}_{100}^{(b)}$ is the set of concatenated sample pairs for sample $b$. 

\begin{equation}
\label{eqn:localloss}
\mathcal{L}_{ll} = \frac{1}{N_{\text{total}}} \sum_{b=1}^{B} \sum_{(m_1, m_2) \in \mathcal{R}_{100}^{(b)}} \left\| F\left( \left[ \mathbf{e}_{m_1}^{(b)};\ \mathbf{e}_{m_2}^{(b)} \right] \right) - \Delta_{m_1,m_2}^{(b)} \right\|_2^2
\end{equation}

\subsection{Combined Loss}
Our combined loss is a weighted sum of $L_\text{jepa}$ (Eq. \ref{eq:jepa_loss}) and $L_{ll}$ (Eq. \ref{eqn:localloss}), with $\lambda$ denoted as weight, shown in Eq. \ref{eqn:Combined_loss}. We ablate this $\lambda$ weighting in Table \ref{tab:Ablation}.

\begin{align}
\label{eqn:Combined_loss}
\mathcal{L}_{\text{combined}} &= \lambda \cdot \mathcal{L}_{\text{jepa}} + (1 - \lambda) \cdot \mathcal{L}_{ll}
\end{align}

\section{Experiments}
All experiments were performed on the publicly available CAMUS dataset \cite{leclerc_deep_2019}. CAMUS is a cardiac ultrasound dataset containing clinical exams from 500 patients from University Hospital of St Etienne, France. Data was collected using a GE Vivid E95 ultrasound scanner and GE M5S probe. 2D apical four chamber and two chamber view sequence are provided for each patient containing at least one full cardiac cycle. Annotations for left ventricle endocardium (LV Endo), left ventricle Epicardium (LV Epi) and left atrium wall (LA wall) are provided.  

\subsection{Experimental Setup}
All experiments were implemented using Pytorch and performed on NVIDIA L40S 48GB GPUs. We used a batch-size of $4$ during pre-training and downstream training, with $16$ frames sampled per video. We used a frame step of $4$ and spatial resolution of $224 \times 224$ pixels. Published pre-trained weights for both V-JEPA and VideoMAE ViT-L models were used before pre-training on CAMUS dataset. For downstream training after pre-training we freeze the ViT-L encoder, use attentive probing, before passing to a shallow decoder consisting of $2$ transpose convolutional layers. This evaluation is similar to described in the V-JEPA paper \cite{bardes_adrien_revisiting_2024}, but with a shallow decoder used to obtain a segmentation output. Pre-training was run for $300$ epochs using AdamW with a $20$-epoch warmup and a cosine learning rate schedule from $0.0002$ to $1e^{-6}$. Downstream training was also run for $300$ epochs using AdamW, cross-entropy loss, and a cosine learning rate schedule from $1e^{-3}$ to $0$.

\subsection{Evaluation Metrics}
To evaluate the performance of our method, we used: Dice Similarity Coefficient (DSC $= \frac{2 \cdot |y_{\text{pred}} \cap y_{\text{true}}|}{|y_{\text{pred}}| + |y_{\text{true}}|}$), Jaccard Index (JI $= \frac{|y_{\text{pred}} \cap y_{\text{true}}|}{|y_{\text{pred}} \cup y_{\text{true}}|}$), precision (PPV $= \frac{|y_{\text{pred}} \cap y_{\text{true}}|}{|y_{\text{pred}}|}$) and recall (Rec. $= \frac{|y_{\text{pred}} \cap y_{\text{true}}|}{|y_{\text{true}}|}$). $y_{\text{pred}}$ and $y_{\text{true}}$ represent the predicted segmentation mask and ground truth segmentation masks, respectively.

\section{Results and Discussion}
We compare segmentation performance between a supervised and pre-trained ViT-L model. We show results for pre-trained methods: VideoMAE, V-JEPA and V-JEPA with our localisation auxiliary task added. With ViT-L/16 the smallest V-JEPA model available with published weights, we demonstrate the impact when 12 and 16 transformer blocks are frozen during pre-training.

\subsection{Ablation Results}
We include an ablation study to investigate the impact of $\lambda$ weighting our combined loss function (Eq. \ref{eqn:Combined_loss}). Table \ref{tab:Ablation} indicates an optimal $\lambda$ weighting of $0.25$, showing validation set performance using 100\% training samples. We found this weighting remains optimal across 10\%, 20\% and 50\% subsets.

\begin{table}[t!]
\centering
\caption{Effect of varying $\lambda$ on validation set. DSC results are shown.}
\label{tab:Ablation}
\begin{tabular}{c|cccc}
\hline
\multirow{2}{*}{\textbf{Method}} & \multicolumn{4}{c}{\textbf{$\lambda$ Setting}}                                                       \\ \cline{2-5} 
                                 & \multicolumn{1}{c|}{0.9}   & \multicolumn{1}{c|}{0.75}  & \multicolumn{1}{c|}{0.5}   & 0.25           \\ \hline
V-Jepa + LL                      & \multicolumn{1}{c|}{0.696} & \multicolumn{1}{c|}{0.658} & \multicolumn{1}{c|}{0.708} & \textbf{0.754} \\ \hline
V-Jepa (12b) + LL                & \multicolumn{1}{c|}{0.782} & \multicolumn{1}{c|}{0.787} & \multicolumn{1}{c|}{0.780} & \textbf{0.805} \\ \hline
V-Jepa (16b) + LL                & \multicolumn{1}{c|}{0.810} & \multicolumn{1}{c|}{0.812} & \multicolumn{1}{c|}{0.817} & \textbf{0.818} \\ \hline
\end{tabular}
\end{table}

\subsection{Quantitative Results}
Our results in Table \ref{tab:Overall_results} show V-JEPA pre-training improves downstream segmentation performance in the CAMUS dataset with all V-JEPA variants outperforming both VideoMAE and the supervised only ViT-L baselines. We show a 10.8\% and 14\% improvement in DSC by increasing the number of frozen transformer blocks using 100\% and 10\% training samples respectively. Freezing ViT-L transformer blocks during pre-training reduces over-fitting to the CAMUS dataset by limiting trainable parameters, while leveraging publicly available pre-trained weights for adaptation to our US domain. Furthermore we demonstrate improved performance using our local loss in V-JEPA pre-training. Adding local loss improved DSC using 100\% training samples by 1.07\% ($p = 1.5e^{-2}$), 3.40\% ($p = 2e^{-19}$) and 0.7\% ($p = 4.4e^{-3}$) for V-JEPA, V-JEPA (12b) and V-JEPA (16b) configurations respectively. These improvements are all statistically significant with $p$-values < 0.05. Similarly with 10\% training samples we show significant segmentation performance improvement of 7.45\% ($p = 2.2e^{-21}$), 8.35\% ($p = 2.9e^{-31}$) and 2.31\% ($p = 3.2e^{-8}$) for V-JEPA, V-JEPA (12b) and V-JEPA (16b) configurations respectively, with $p$-values < 0.05. Furthermore Table \ref{tab:Overall_results} shows similar improvement in JI results when local loss is added. When inspecting PPV and recall, we see greater improvement to recall with local loss added, particularly as training samples become more limited. 

These results demonstrate that adding our local loss auxiliary task benefits V-JEPA pre-training, enhancing representation quality on the small CAMUS dataset, particularly benefitting downstream segmentation performance under limited data scenarios, i.e. with 10\% training samples. 

\begin{table}[t!]
\centering
\caption{Quantitative comparison of US video segmentation on CAMUS dataset for different \% of training samples. The overall results across all videos in the test set are presented. All models use the ViT-L/16 encoder. SD indicates the standard deviation.}
\label{tab:Overall_results}
\scalebox{0.85}{
\begin{tabular}{c|c|c|c|c|c}
\Xhline{1.5pt}
\textbf{Method} &
  \textbf{\begin{tabular}[c]{@{}c@{}}\% training \\ samples\end{tabular}} &
  \textbf{DSC $\pm$ SD} &
  \textbf{JI $\pm$ SD} &
  \textbf{PPV $\pm$ SD} &
  \textbf{Recall $\pm$ SD} \\ \Xhline{1pt}
\multirow{4}{*}{Supervised ViT-L} &
  100 &
  0.681 ± 0.076 &
  0.641 ± 0.116 &
  0.794 ± 0.123 &
  0.746 ± 0.117 \\ \cline{2-6} 
                                   & 50  & 0.665 ± 0.080          & 0.607 ± 0.114          & 0.766 ± 0.131          & 0.715 ± 0.119          \\ \cline{2-6} 
                                   & 20  & 0.635 ± 0.089          & 0.571 ± 0.136          & 0.740 ± 0.148          & 0.679 ± 0.146          \\ \cline{2-6} 
                                   & 10  & 0.605 ± 0.089          & 0.551 ± 0.138          & 0.724 ± 0.162          & 0.653 ± 0.152          \\ \hline
\multirow{4}{*}{VideoMAE}            & 100 & 0.708 ± 0.076                     & 0.665 ± 0.103                     & 0.797 ± 0.113                     & 0.783 ± 0.112                      \\ \cline{2-6} 
                                   & 50  & 0.652 ± 0.077                     & 0.615 ± 0.117                      & 0.762 ± 0.133                     & 0.738 ± 0.128                     \\ \cline{2-6} 
                                   & 20  & 0.643 ± 0.073                     & 0.603 ± 0.119                     & 0.749 ± 0.138                      & 0.725 ± 0.130                     \\ \cline{2-6} 
                                   & 10  & 0.590 ± 0.080                      & 0.560 ± 0.138                     & 0.714 ± 0.160                      & 0.677 ± 0.155                      \\ \Xhline{1.5pt}
\multirow{4}{*}{V-JEPA}            & 100 & 0.747 ± 0.067          & 0.679 ± 0.093          & 0.814 ± 0.100          & 0.788 ± 0.094          \\ \cline{2-6} 
                                   & 50  & 0.729 ± 0.069          & 0.669 ± 0.097          & 0.799 ± 0.104          & 0.788 ± 0.099          \\ \cline{2-6} 
                                   & 20  & 0.681 ± 0.088          & 0.608 ± 0.117          & 0.753 ± 0.134          & 0.727 ± 0.123          \\ \cline{2-6} 
                                   & 10  & 0.644 ± 0.094          & 0.574 ± 0.125          & 0.721 ± 0.147          & 0.698 ± 0.142          \\ \hline
\multirow{4}{*}{\begin{tabular}[c]{@{}c@{}}V-Jepa + LL\\ (ours)\end{tabular}} & 
100 & 
0.755 ± 0.075                      & 0.674 ± 0.097                      & 0.800 ± 0.107                      & 0.793 ± 0.095                      \\ \cline{2-6} 
                                   & 50  & 0.747 ± 0.078                     & 0.659 ± 0.097                      & 0.761 ± 0.111                      & 0.814 ± 0.093                      \\ \cline{2-6} 
                                   & 20  & 0.678 ± 0.081                      & 0.593 ± 0.103                     & 0.757 ± 0.127                     & 0.714 ± 0.115                     \\ \cline{2-6} 
                                   & 10  & 0.692 ± 0.092                     & 0.600 ± 0.110                     & 0.745 ± 0.130                     & 0.729 ± 0.123                     \\ \Xhline{1.5pt}
\multirow{4}{*}{V-JEPA (12b)}      & 100 & 0.795 ± 0.053          & 0.736 ± 0.074          & 0.862 ± 0.077          & 0.826 ± 0.072          \\ \cline{2-6} 
                                   & 50  & 0.775 ± 0.058          & 0.705 ± 0.077          & 0.846 ± 0.085          & 0.797 ± 0.078          \\ \cline{2-6} 
                                   & 20  & 0.751 ± 0.035          & 0.678 ± 0.040          & 0.810 ± 0.050          & 0.792 ± 0.053          \\ \cline{2-6} 
                                   & 10  & 0.683 ± 0.082          & 0.609 ± 0.107          & 0.756 ± 0.125          & 0.731 ± 0.118          \\ \hline
\multirow{4}{*}{\begin{tabular}[c]{@{}c@{}}V-Jepa (12b) + LL\\ (ours)\end{tabular}} & 100 & 0.822 ± 0.048          & 0.761 ± 0.072          & 0.874 ± 0.069          & 0.850 ± 0.068          \\ \cline{2-6} 
                                   & 50  & 0.804 ± 0.050          & 0.743 ± 0.074          & 0.856 ± 0.076          & 0.845 ± 0.070          \\ \cline{2-6} 
                                   & 20  & 0.774 ± 0.062          & 0.706 ± 0.089          & 0.829 ± 0.093          & 0.819 ± 0.091          \\ \cline{2-6} 
                                   & 10  & 0.740 ± 0.069          & 0.671 ± 0.967          & 0.812 ± 0.101          & 0.782 ± 0.105          \\ \Xhline{1.5pt}
\multirow{4}{*}{V-JEPA (16b)} &
  100 &
  0.828 ± 0.044 &
  \textbf{0.779 ± 0.064} &
  0.880 ± 0.064 &
  \textbf{0.868 ± 0.067} \\ \cline{2-6} 
                                   & 50  & 0.803 ± 0.051          & 0.747 ± 0.073          & 0.861 ± 0.073          & 0.844 ± 0.074          \\ \cline{2-6} 
                                   & 20  & \textbf{0.780 ± 0.068} & \textbf{0.707 ± 0.092} & \textbf{0.834 ± 0.094} & \textbf{0.814 ± 0.089} \\ \cline{2-6} 
                                   & 10  & 0.734 ± 0.079          & 0.661 ± 0.107          & 0.812 ± 0.110          & 0.765 ± 0.117          \\ \hline
\multirow{4}{*}{\begin{tabular}[c]{@{}c@{}}V-Jepa (16b) + LL\\ (ours)\end{tabular}} &
  100 &
  \textbf{0.834 ± 0.046} &
  0.778 ± 0.068 &
  \textbf{0.882 ± 0.063} &
  0.863 ± 0.065 \\ \cline{2-6} 
                                   & 50  & \textbf{0.824 ± 0.046} & \textbf{0.765 ± 0.068} & \textbf{0.870 ± 0.067} & \textbf{0.859 ± 0.068} \\ \cline{2-6} 
                                   & 20  & 0.779 ± 0.071          & 0.704 ± 0.092          & 0.827 ± 0.098          & \textbf{0.814 ± 0.094} \\ \cline{2-6} 
                                   & 
                                   10  & \textbf{0.751 ± 0.077} & \textbf{0.668 ± 0.099} & 0.811 ± 0.104          & \textbf{0.778 ± 0.105} \\ \Xhline{1.5pt}
\end{tabular}
}
\end{table}
\subsection{Qualitative Results}
\label{qual_results}
\begin{figure*}[t!]
    \centering    \includegraphics[width=1\textwidth]{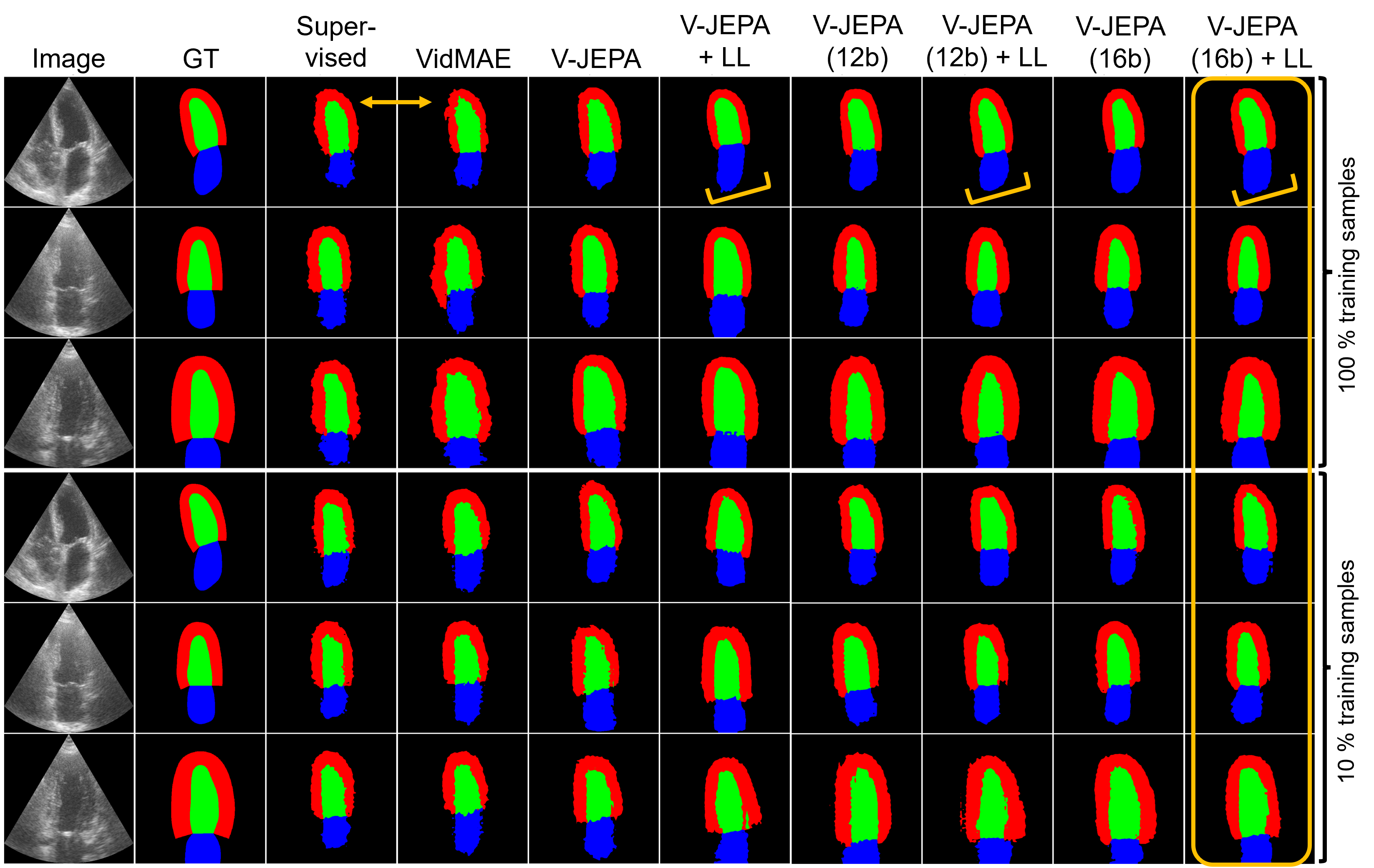}
    \caption{Qualitative evaluation on CAMUS dataset. 3 example videos were chosen at frame 9. Segmentation predictions across all methods are presented, using 100\% and 10\% training samples. LV endocardium, LV epicardium, LA wall are indicated in green, red and blue respectively. Orange annotations highlight key points discussed in the qualitative results section (see \ref{qual_results}).}
    \label{fig:Qual_eval}
\end{figure*}

Segmentation predictions for 3 example frames are shown for each method in Figure \ref{fig:Qual_eval} for both 100\% and 10\% training samples. We can see V-JEPA methods perform better at segmenting each class relative to the ground truth labels. This is demonstrated by smoother class boundaries compared to the supervised ViT and VidMAE methods. The orange double-headed arrow in Figure \ref{fig:Qual_eval} highlights jagged boundaries in these approaches. Using 100\% training samples, adding local loss captures the orientation of the ground truth more effectively compared to each corresponding V-JEPA baseline (highlighted in Figure \ref{fig:Qual_eval}, row 1, with orange brackets). As expected, with much fewer training samples segmentation performance worsens. With 10\% training samples, the orientation of the ground truth mask is not captured well across all segmented predictions (see Figure \ref{fig:Qual_eval}, row 4). Secondly, segmented boundaries lose smoothness compared to the ground truth labels. However, the relative size of the overall segmentations are captured best using VJEPA (12b), VJEPA (12b) + LL, VJEPA (16b) and VJEPA (16b) + LL variants, with VJEPA (16b) + LL showing best segmentation results overall, highlighted in orange in Figure \ref{fig:Qual_eval}.

\section{Conclusions}
In this work we explored the performance of V-JEPA on cardiac ultrasound data. V-JEPA outperformed the commonly used VideoMAE approach for self-supervised learning on video data. However, with these methods well-suited to transformer-based models, performance can suffer on smaller medical datasets. We proposed a 3D relative localisation auxiliary task to improve V-JEPA pre-training when data is limited. This approach strengthens ViT spatial locality understanding, leading to improved representation learning and significantly better downstream segmentation performance. Future work will apply this method to a broader range of ultrasound video datasets, integrating complementary strategies, such as hierarchical transformers, to enhance performance on small datasets further.

\newpage
\bibliographystyle{splncs04}
\bibliography{references}

\begin{thebibliography}{10}
\providecommand{\url}[1]{\texttt{#1}}
\providecommand{\urlprefix}{URL }
\providecommand{\doi}[1]{https://doi.org/#1}

\bibitem{akkaya_enhancing_2024}
Akkaya, I.B., Kathiresan, S.S., Arani, E., Zonooz, B.: Enhancing performance of
  vision transformers on small datasets through local inductive bias
  incorporation. Pattern Recognition  \textbf{153},  110510 (Sep 2024)

\bibitem{bardes_adrien_revisiting_2024}
Bardes, A., Garrido, Q., Ponce, J., Chen, X., Rabbat, M., LeCun, Y., Assran,
  M., Ballas, N.: Revisiting {Feature} {Prediction} for {Learning} {Visual}
  {Representations} from {Video}. arXiv  (2024),
  \url{https://arxiv.org/abs/2404.08471}

\bibitem{brattain_machine_2018}
Brattain, L.J., Telfer, B.A., Dhyani, M., Grajo, J.R., Samir, A.E.: Machine
  learning for medical ultrasound: status, methods, and future opportunities.
  Abdominal Radiology  \textbf{43}(4),  786--799 (Apr 2018)

\bibitem{chen_contrastive_2023}
Chen, L., Rubin, J., Ouyang, J., Balaraju, N., Patil, S., Mehanian, C.,
  Kulhare, S., Millin, R., Gregory, K.W., Gregory, C.R.: Contrastive
  self-supervised learning for spatio-temporal analysis of lung ultrasound
  videos. pp.~1--5. IEEE (2023)

\bibitem{dosovitskiy_image_2020}
Dosovitskiy, A., Beyer, L., Kolesnikov, A., Weissenborn, D., Zhai, X.,
  Unterthiner, T., Dehghani, M., Minderer, M., Heigold, G., Gelly, S.: An
  {Image} is {Worth} 16x16 {Words}: {Transformers} for {Image} {Recognition} at
  {Scale} (2020)

\bibitem{e_lamoureux_segmenting_2023}
{E. Lamoureux}, {S. Ayromlou}, {S. N. Ahmadi Amiri}, {H. Rhodin}: Segmenting
  {Cardiac} {Ultrasound} {Videos} {Using} {Self}-{Supervised} {Learning}. In:
  2023 45th {Annual} {International} {Conference} of the {IEEE} {Engineering}
  in {Medicine} \& {Biology} {Society} ({EMBC}). pp.~1--7 (Jul 2023)

\bibitem{ellis_self-supervised_2025}
Ellis, E., Bulpitt, A., Parsa, N., Byrne, M.F., Ali, S.: A {Self}-{Supervised}
  {Framework} for {Improved} {Generalisability} in {Ultrasound} {B}-mode
  {Image} {Segmentation}. arXiv preprint arXiv:2502.02489  (2025)

\bibitem{fu_anatomy-aware_2022}
Fu, Z., Jiao, J., Yasrab, R., Drukker, L., Papageorghiou, A.T., Noble, J.A.:
  Anatomy-{Aware} {Contrastive} {Representation} {Learning} for {Fetal}
  {Ultrasound}. Computer vision - ECCV. European Conference on Computer Vision:
  proceedings. European Conference on Computer Vision  \textbf{2022},  422--436
  (Oct 2022)

\bibitem{jiao_self-supervised_2020}
Jiao, J., Droste, R., Drukker, L., Papageorghiou, A.T., Noble, J.A.:
  Self-{Supervised} {Representation} {Learning} for {Ultrasound} {Video}.
  Proceedings. IEEE International Symposium on Biomedical Imaging
  \textbf{2020},  1847--1850 (Apr 2020)

\bibitem{jiao_usfm_2024}
Jiao, J., Zhou, J., Li, X., Xia, M., Huang, Y., Huang, L., Wang, N., Zhang, X.,
  Zhou, S., Wang, Y.: Usfm: {A} universal ultrasound foundation model
  generalized to tasks and organs towards label efficient image analysis.
  Medical Image Analysis  \textbf{96},  103202 (2024)

\bibitem{leclerc_deep_2019}
Leclerc, S., Smistad, E., Pedrosa, J., Østvik, A., Cervenansky, F., Espinosa,
  F., Espeland, T., Berg, E.A.R., Jodoin, P.M., Grenier, T., Lartizien, C.,
  D’hooge, J., Lovstakken, L., Bernard, O.: Deep {Learning} for
  {Segmentation} {Using} an {Open} {Large}-{Scale} {Dataset} in {2D}
  {Echocardiography}. IEEE Transactions on Medical Imaging  \textbf{38}(9),
  2198--2210 (2019)

\bibitem{lecun_path_2022}
LeCun, Y.: A path towards autonomous machine intelligence version 0.9. 2,
  2022-06-27. Open Review  \textbf{62}(1),  1--62 (2022)

\bibitem{liu_self-supervised_2023}
Liu, X., Zhang, F., Hou, Z., Mian, L., Wang, Z., Zhang, J., Tang, J.:
  Self-{Supervised} {Learning}: {Generative} or {Contrastive}. IEEE
  Transactions on Knowledge and Data Engineering  \textbf{35}(1),  857--876
  (2023)

\bibitem{liu_efficient_2021}
Liu, Y., Sangineto, E., Bi, W., Sebe, N., Lepri, B., Nadai, M.: Efficient
  training of visual transformers with small datasets. Advances in Neural
  Information Processing Systems  \textbf{34},  23818--23830 (2021)

\bibitem{liu_swin_2021}
Liu, Z., Lin, Y., Cao, Y., Hu, H., Wei, Y., Zhang, Z., Lin, S., Guo, B.: Swin
  transformer: {Hierarchical} vision transformer using shifted windows. pp.
  10012--10022 (2021)

\bibitem{quien_ultrasound_2018}
Quien, M.M., Saric, M.: Ultrasound imaging artifacts: {How} to recognize them
  and how to avoid them. Echocardiography  \textbf{35}(9),  1388--1401 (2018)

\bibitem{szijarto_masked_2024}
Szijártó, A., Magyar, B., Szeier, T.A., Tolvaj, M., Fábián, A., Lakatos,
  B.K., Ladányi, Z., Bagyura, Z., Merkely, B., Kovács, A.: Masked
  {Autoencoders} for {Medical} {Ultrasound} {Videos} {Using} {ROI}-{Aware}
  {Masking}. pp. 167--176. Springer (2024)

\bibitem{tong_videomae_2022}
Tong, Z., Song, Y., Wang, J., Wang, L.: {VideoMAE}: masked autoencoders are
  data-efficient learners for self-supervised video pre-training. In:
  Proceedings of the 36th {International} {Conference} on {Neural}
  {Information} {Processing} {Systems}. {NIPS} '22, Red Hook, NY, USA (2022)

\bibitem{wang_pyramid_2021}
Wang, W., Xie, E., Li, X., Fan, D.P., Song, K., Liang, D., Lu, T., Luo, P.,
  Shao, L.: Pyramid vision transformer: {A} versatile backbone for dense
  prediction without convolutions. pp. 568--578 (2021)

\bibitem{gomez_fetal_2025}
Zhang, K., Jiao, J., Noble, J.A.: Fetal {Ultrasound} {Video} {Representation}
  {Learning} {Using} {Contrastive} {Rubik}’s {Cube} {Recovery}. In:
  Simplifying {Medical} {Ultrasound}, vol. 15186, pp. 187--197. Springer Nature
  Switzerland, Cham (2025)

\bibitem{zhu_understanding_2023}
Zhu, H., Chen, B., Yang, C.: Understanding why vit trains badly on small
  datasets: {An} intuitive perspective. arXiv preprint arXiv:2302.03751  (2023)

\end{thebibliography}
\end{document}